*Research Article*

# Recognizing Uncertainty in Speech

**Heather Pon-Barry and Stuart M. Shieber**

*School of Engineering and Applied Sciences, Harvard University, 33 Oxford Street, Cambridge, MA 02138, USA*

Correspondence should be addressed to Heather Pon-Barry, ponbarry@eecs.harvard.edu

We address the problem of inferring a speaker's level of certainty based on prosodic information in the speech signal, which has application in speech-based dialogue systems. We show that using phrase-level prosodic features centered around the phrases causing uncertainty, in addition to utterance-level prosodic features, improves our model's level of certainty classification. In addition, our models can be used to predict which phrase a person is uncertain about. These results rely on a novel method for eliciting utterances of varying levels of certainty that allows us to compare the utility of contextually-based feature sets. We elicit level of certainty ratings from both the speakers themselves and a panel of listeners, finding that there is often a mismatch between speakers' internal states and their perceived states, and highlighting the importance of this distinction.

## 1. Introduction

Speech-based technology has become a familiar part of our everyday lives. Yet, while most people can think of an instance where they have interacted with a call-center dialogue system, or command-based smartphone application, few would argue that the experience was as natural or as efficient as conversing with another human. To build computer systems that can communicate with humans using natural language, we need to know more than just the words a person is saying; we need to have an understanding of his or her internal mental state.

Level of certainty is an important component of internal state. When people are conversing face to face, listeners are able to sense whether the speaker is certain or uncertain through contextual, visual, and auditory cues [1]. If we enable computers to do the same, we can improve how applications such as spoken tutorial dialogue systems [2], language learning systems [3], and voice search applications [4] interact with users.

Although humans can convey their level of certainty through audio and visual channels, we focus on the audio (the speaker's prosody) because in many potential applications, there is audio input but no visual input. On tasks ranging from detecting frustration [5] to detecting flirtation [6], prosody has been shown to convey information about a speaker's emotional and mental state [7] and about their social intentions. Our work builds upon this, as well as a small body of work on identifying prosodic cues to level of certainty [1] and classifying a speaker's certainty [8]. The intended application of such work is for dialogue systems to appropriately respond to a speaker based on their level of certainty as exposed in their prosody, for example, by altering the content of system responses [9] or by altering the emotional coloring of system responses [10].

Our primary goal is to determine whether prosodic information from a spoken utterance can be used to determine how certain a speaker is. We argue that speech-based applications will benefit from knowing the speaker's level of certainty. But "level of certainty" has multiple interpretations. It may refer to how certain a person sounds, the *perceived* level of certainty. This definition is reasonable because we are looking for prosodic cues—we want our system to hear whatever it is that humans hear. Not surprisingly, this is the definition that has been assumed in previous work on classifying level of certainty [8]. However, in applications such as spoken tutoring systems [9] and second language learning systems [3], we would like to know how certain speakers actually are—their *internal* level of certainty, in addition to how certain they are perceived to be. This knowledge affects the inferences such systems can make about the speaker's internal state, for example, whether the speaker has a misconception, makes a lucky guess, or might benefit from some encouragement. Getting a ground truth measurement of a speaker's internal level of certainty is nearly impossible, though by asking speakers to rate their own level of certainty, we can get a good approximation, the *self-reported* level of certainty, which we use as a proxy for internal level of certainty in this paper.

In the past work on using prosody to classify level of certainty, no one has attempted to classify a person's internal level of certainty. Therefore, one novel contribution of our work is that we collect self-reported level of certainty assessments from the speakers, in addition to collecting perceived level of certainty judgements from a set of listeners. We look at whether simple machine learning models can classify self-reported level of certainty based on the prosody of an utterance. We also show that knowing the utterance's perceived level of certainty helps make more accurate predictions about the self-reported level of certainty.

Returning to the problem of classifying perceived level of certainty, we present a basic model that uses prosodic information to classify utterances as certain, uncertain, or neutral. This model performs better than a trivial baseline model (choosing the most common class), corroborating results of prior work, but we also show for the first time that the prosody is crucial in achieving this performance by comparing to a substantive nonprosodic baseline.

In some applications, for instance, language learning and other tutorial systems, we have information as to which phrase in an utterance is the probable source of uncertainty. We ask whether we can improve upon the basic model by taking advantage of this information. We show that the prosody of this phrase and of its surrounding regions help make better certainty classifications. Conversely, we show that our models can be used to make an informed guess about which phrase a person is uncertain about when we do not know which phrase is the probable source of uncertainty.

Because existing speech corpora are not sufficient for answering such questions, we designed a novel method for eliciting utterances of varying levels of certainty. Our corpus contains sets of utterances that are lexically identical but differ in their level of certainty; thus, any differences in prosody can be attributed to the speaker's level of certainty. Further, we control which words or phrases within an utterance are responsible for variations in the speaker's level of certainty. We collect level of certainty self-reports from the speakers and perceived level of certainty ratings from five human judges. This corpus enables us to address the questions above.

The four main contributions of this work are

(i) a methodology for collecting uncertainty data, plus an annotated corpus;

(ii) an examination of the differences between perceived uncertainty and self-reported uncertainty;

(iii) corroboration and extension of previous results in predicting perceived uncertainty;

(iv) a technique for computing prosodic features from utterance segments that both improves uncertainty classification and can be used to determine the cause of uncertainty.

Our data collection methodology is described in Section 2. We find that perceived certainty accurately reflects self-reported certainty for only half of the utterances in our corpus. In Section 3, we discuss this difference and highlight the importance of collecting both quantities. We then present a model for classifying a person's *self-reported certainty* in Section 4. In Section 5, we describe a basic classifier that uses prosodic features computed at the utterance level to classify how certain a person is *perceived* with an accuracy of 69%. The performance of the basic classifier compares well to prior work [8]. We go beyond prior work by showing improvement over a nonprosodic baseline. We improve upon this basic model by identifying salient prosodic features from utterance segments that correspond to the probable source of uncertainty. We show, in Section 6, that models trained on such features reach a classification accuracy of 75%. Lastly, in Section 7, we explain how the models from Section 6 can be used to determine which of two phrases within an utterance is the cause of a speaker's uncertainty with an accuracy of over 90%.

## 2. Methodology for Creating an Uncertainty Corpus

Our results are enabled by a data collection method that is motivated by four main criteria.

(1) For each speaker, we want to elicit utterances of varying levels of certainty.

(2) We want to isolate the words or phrases within an utterance that could cause the speaker to be uncertain.

(3) To ensure that differences in prosody are not due to the particular phonemes in the words or the number of words in the sentence, we want to collect utterances across speakers that are lexically similar.

(4) We want the corpus to contain multiple instances of the same word or phrase in different contexts.

Prior work on certainty prediction used spontaneous speech in the context of a human-computer dialogue system. Such a corpus cannot be carefully controlled to satisfy these criteria. For this reason, we developed a novel data collection method based on nonspontaneous read speech with speaker options.

Because we want consistency across utterances (criterion 3), we collect nonspontaneous as opposed to spontaneous speech. Although spontaneous speech is more natural, we found in pilot experiments that the same set of acoustic features were significantly correlated with perceived level of certainty in both spontaneous speech and nonspontaneous speech conditions. To ensure varying levels of certainty (criterion 1), we could not have speakers just read a given sentence. Instead, the speakers are given multiple options of what to read and thus are forced to make a decision. Because we want to isolate the phrases causing uncertainty (criterion 2), the multiple options to choose among occur at the word or phrase level, and the rest of the sentence is fixed. Consider the example below, in the domain of answering questions about using public transportation in Boston.

> Q: How can I get from Harvard to the Silver Line?
>
> A: Take the red line to ____.
>
> (a) South Station
> (b) Downtown Crossing

In this example, the experimenter first asks a question aloud, *How can I get from Harvard to the Silver Line?* Without seeing the options for filling in the slot, the speakers see the fixed part of the response, *Take the red line to ____*, which we refer to as the *context*. They have unlimited time to read over the context. Upon a keypress, *South Station* and *Downtown Crossing*, which we refer to as the *target words*, are displayed below the context. Speakers are instructed to choose the best answer and read the full sentence aloud upon hearing a beep, which is played 1.5 seconds after the target words appear. This forces them to make their decisions quickly. Because the speakers have unlimited time to read over the context before seeing the target words, the target word corresponds to the decision the speakers have to make, and we consider it to be the *source* of the uncertainty. In this way, we are able to isolate the phrases causing uncertainty (criterion 2).

To elicit both certain and uncertain utterances from each speaker (criterion 1), the transit questions differ in the amount of real-world knowledge needed to answer the question correctly. Some of the hardest items contain two or three slots to be filled. Because we want the corpus to contain multiple instances of the same word in different contexts (criterion 4), the potential target words are repeated throughout the experiment. This allows us to see whether individual speakers have systematic ways of conveying their level of certainty.

In addition to the public transportation utterances, we elicited utterances in a second domain: choosing vocabulary words to complete a sentence. An example item is shown below.

> Only the ____ workers in the office laughed at all of the manager's bad jokes.
>
> (a) pugnacious
> (b) craven
> (c) sycophantic
> (d) spoffish

In the vocabulary domain, speakers are instructed to choose the word that best completes the sentence. To ensure that even the most well-read participants would be uncertain at times (criterion 1), the potential target words include three extremely infrequent words (e.g., *spoffish*), and in five of the 20 items, *none* of the potential target words fit well in the context, generating further speaker uncertainty.

The corpus contains 10 items in the transit domain and 20 items in the vocabulary domain, each spoken by 20 adult native English speakers, for a total of 600 utterances. The mean and standard deviation of the age of the speakers was $22.35 \pm 3.13$. After each utterance, the speakers rated their own level of certainty on a 5-point scale, where 1 is labeled as "very uncertain" and 5 is labeled as "very certain." We will refer to this rating as the "*self-reported level of certainty.*" As we show in the next section by examining these self-reports of certainty, our data collection methodology fulfills the crucial criterion (1) of generating a broad range of certainty levels.

In addition, five human judges listened to the utterances and judged how certain the speaker sounded, using the same 5-point scale (where 1 is labeled as "very uncertain" and 5 is labeled as "very certain"). The mean and standard deviation of the age of the listeners was $21.20 \pm 0.84$. The listeners did not have any background in linguistics or speech annotation. They listened to the utterances in a random order and had no knowledge of the target words, the questions for the transit items, or the instructions that the speakers were given. The average interannotator agreement (Kappa) was 0.45, which is on par with past work in emotion detection [7, 8]. We refer to the mean of the five listeners' ratings for an utterance as the "*perceived level of certainty.*"

The data collection materials, level of certainty annotations, and prosodic and nonprosodic feature values for this corpus will be made available through the Dataverse Network: http://dvn.iq.harvard.edu/dvn/dv/ponbarry/.

## 3. Self-Reported versus Perceived Level of Certainty

Since we elicit both self-reported and perceived level of certainty judgments, we are able to assess whether perceived level of certainty is an accurate reflection of a person's internal level of certainty. In our corpus, we find that this is not the case. As illustrated in Figure 1, the distribution of self-ratings is more heavily concentrated on the uncertain side (mean $2.6 \pm 1.4$), whereas the annotators' ratings are more heavily concentrated on the certain side (mean $3.5 \pm 1.1$). Correlation between the two measures of uncertainty is 0.42. Furthermore, the heat map in Figure 1 demonstrates that this discrepancy is not random; the concentration of darker squares above the diagonal shows that listeners rated speakers as being more certain than they actually were more often than the reverse case. Of the 600 utterances, 41% had perceived ratings that were more than one unit *greater than* the self-reported rating and only 8% had perceived ratings were more than one unit *less than* the self-reported rating. Thus, perceived level of certainty is not an ideal measure of the self-reports, our proxy for internal level of certainty.

Previous work on level of certainty classification has focused on classifying an utterance's *perceived* level of certainty. However, in many applications such as spoken tutoring systems [9] and second language learning systems [3], we would like to know how certain speakers *actually are* in addition to how certain they are perceived to be. To illustrate why it is important to have both measures of certainty, we define two new categories pertaining to level of certainty: *self-awareness* and *transparency*. Knowing whether speakers are self-aware or whether they are transparent affects the inferences speech systems can make about speakers' internal states, for example, whether they have a misconception, make a lucky guess, or might benefit from some encouragement.

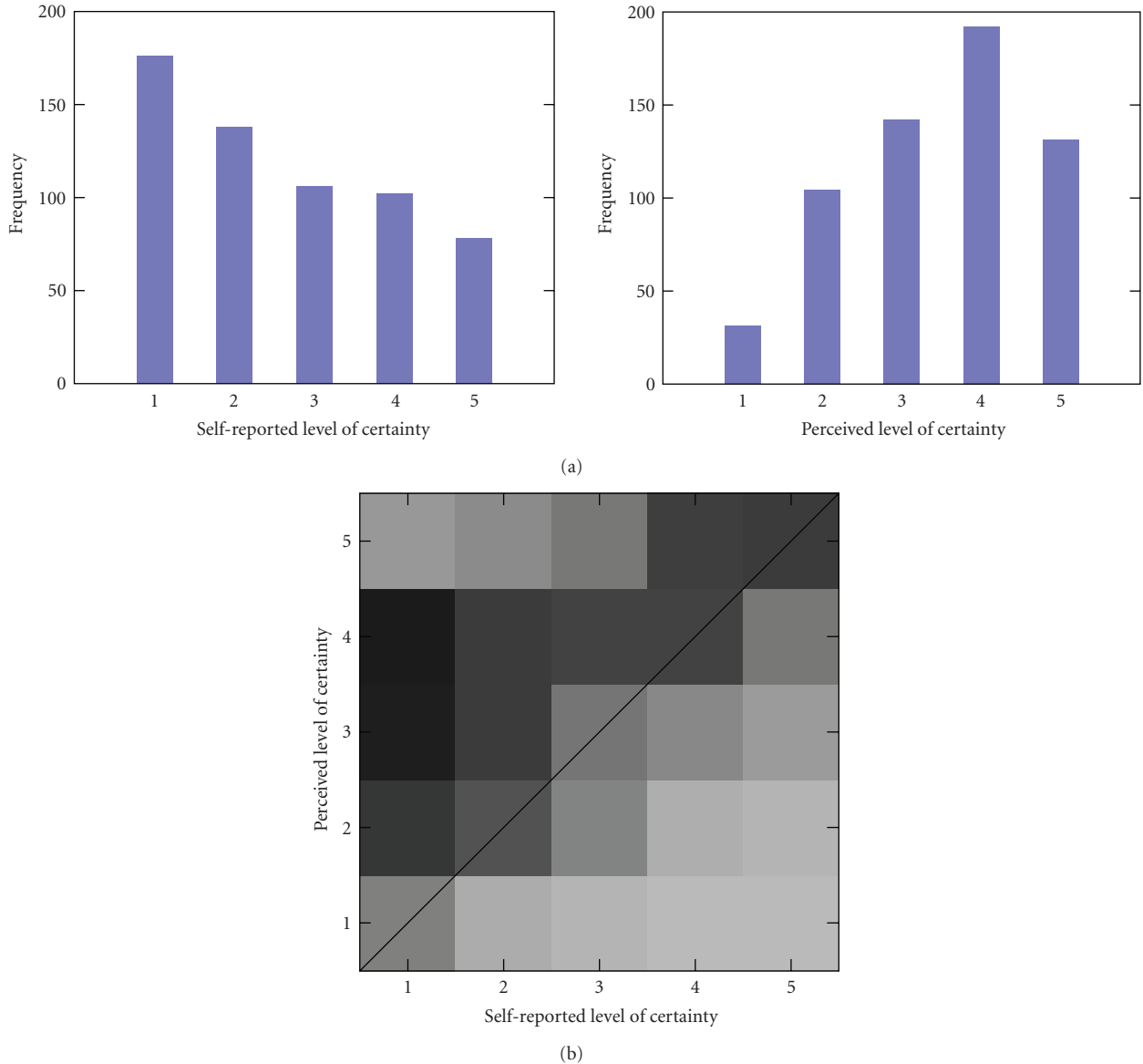

Figure 1: (a) Histograms illustrating the distribution of self-reported certainty and (quantized) perceived certainty in our corpus; (b) heat map illustrating the relative frequencies of utterances grouped according to both self-reported certainty and (quantized) perceived certainty (darker means more frequent).

*3.1. Self-Awareness.* The concept of self-awareness applies to utterances whose correctness can be determined. We consider speakers to be *self-aware* if they feel certain when correct and feel uncertain when incorrect. The four possible combinations of correctness versus internal level of certainty are illustrated in Figure 2. Self-awareness is similar (though not identical) to the "feeling of knowing" measure of Smith and Clark [11]. In conversational, question-answering settings, speakers systematically convey their feeling of knowing through both auditory and visual prosodic cues [12].

For educational applications, systems that can assess self-awareness can assess whether or not the user is at a learning impasse [9]. We claim that the most serious learning impasses correspond to the cases where a speaker is *not* self-aware. If a speaker feels certain and is incorrect, then it is likely that they have some kind of *misconception*. If a speaker feels uncertain and is correct, they either *lack confidence* or made a lucky guess. A followup question could be asked by the system to determine whether or not the user made a lucky guess.

|  | Correctness | |
|---|---|---|
|  | Incorrect | Correct |
| Self  UNC | Self-aware | Non-self-aware (lacks confidence or lucky guess) |
| Self  CER | Non-self-aware (misconception) | Self-aware |

FIGURE 2: Self-awareness: we consider speakers to be self-aware if their internal level of certainty reflects the correctness of their utterance.

|  | Perceived | |
|---|---|---|
|  | UNC | CER |
| Self  UNC | Transparent | Opaque (broadcaster) |
| Self  CER | Opaque (meek speaker) | Transparent |

FIGURE 3: Transparency: we consider speakers to be transparent if their internal level of certainty reflects their perceived level of certainty.

For these purposes, we require a binary classification of the levels of certainty and correctness. For both self-reported rating and perceived rating, we map values less than 3 to "uncertain" and values greater than or equal to 3 to "certain." To compute correctness, we code each multiple choice answer or answer tuple as "incorrect" or "correct." Based on this encoding, in our corpus, speakers were self-aware for 73% of the utterances.

*3.2. Transparency.* The concept of speaker *transparency* is independent of an utterance's correctness. We consider speakers to be transparent if they are perceived as certain when they feel certain and are perceived as uncertain when they feel uncertain. The four possible combinations of perceived versus internal level of certainty are illustrated in Figure 3. If a system uses perceived level of certainty to determine what kind of feedback to give the user, then it will give inappropriate feedback to users who are *not* transparent. In our corpus, speakers were transparent in 64% of the utterances. We observed that some speakers acted like radio broadcasters; they sounded very certain even when they felt uncertain. Other speakers had very meek manners of speaking and were perceived as uncertain despite feeling certain. While some speakers consistently fell into one of these categories, others had mixed degrees of transparency. We believe there are many factors that can affect how transparent a speaker is, for example, related work in psychology argues that speakers' beliefs about their transparency and thus the emotions they convey are highly dependent on the context of the interaction [13].

A concept closely related to transparency is the "feeling of another's knowing" [14]—a listener's perception of a speaker's feeling of knowing [11]. This is especially relevant because recent work indicates that spoken tutorial dialogue systems can predict student learning gains better by monitoring the feeling of another's knowing than by monitoring only the correctness of the student's answers [15].

*3.3. Summary.* Our corpus demonstrates that there are systematic differences between perceived certainty and self-reported certainty. Research that treats them as equivalent quantities may be overlooking significant issues. By considering the concepts of self-awareness and transparency, we see how a speech-based system that can estimate both the speaker's perceived and self-reported levels of certainty could make nuanced inferences about the speaker's internal state.

## 4. Modeling Self-Reported Level of Certainty

The ability to sense when speakers are or are not self-aware or transparent allows dialogue systems to give more appropriate feedback. In order to make inferences about self-awareness and transparency, we need to model speakers' internal level of certainty. As stated before, getting a measurement of internal certainty is nearly impossible, so we use self-reported certainty as an approximation. An intriguing possibility is to use information gleaned from perceived level of certainty to more accurately model the self-reported level. This idea bears promise especially given the potential, pursued by ourselves (Section 5) and others [8], of inferring the perceived level of certainty directly from prosodic information. We pursue this idea in this section, showing that a kind of triage on the perceived level of certainty can improve self-report predictions.

*4.1. Prosodic Features.* The prosodic features we use as input in this experiment and reference throughout the paper are listed in Table 1. This set of features was selected in order to be comparable with Liscombe et al. [8], who use these same prosodic features plus dialogue turn-related features in their work on classifying level of certainty. Other recent work on classifying level of certainty uses similar pitch and energy features, plus a few additional f0 features to better approximate the pitch contour, in addition to nonprosodic word-position features [16]. Related research on classifying positive and negative emotion in speech uses a similar set of prosodic features, with the addition of formant-related features, in conjunction with nonprosodic lexical and discourse features [7].

We use WaveSurfer (http://www.speech.kth.se/wavesurfer/) and Praat (http://www.fon.hum.uva.nl/praat/) to compute the feature values. The pitch and intensity features are represented as *z*-scores normalized by speaker; the temporal features are not normalized. The f0 contour is extracted using WaveSurfer's ESPS method. We compute speaking rate as number of syllables divided by speaking duration.

*4.2. Constructing a Model for Self-Reported Certainty.* We build C4.5 decision tree models, using the Weka

TABLE 1: In our experiments, we use a standard set of pitch, intensity, and temporal prosodic features.

| Pitch | min f0 | relative position min f0 |
|---|---|---|
| | max f0 | relative position max f0 |
| | mean f0 | absolute slope (Hz) |
| | stdev f0 | absolute slope (Semi) |
| | range f0 | |
| Intensity | min RMS | relative position min RMS |
| | max RMS | relative position max RMS |
| | mean RMS | stdev RMS |
| Temporal | total silence | percent silence |
| | total duration | speaking duration |
| | speaking rate | |

TABLE 2: Accuracies for classifying self-reported level of certainty for the initial prosody decision tree model and two baselines. The prosody decision tree model is better than choosing the majority class and better than assigning the level to be the same as the perceived level.

| Model | Accuracy |
|---|---|
| Baseline 1: majority class | 52.30 |
| Baseline 2: assign perceived level | 63.67 |
| Single prosody decision tree | 66.33 |

(http://www.cs.waikato.ac.nz/ml/weka/) toolkit to classify self-reported level of certainty based on an utterance's prosody. We code the perceived and self-reported levels of certainty and correctness as binary features as per Section 3.1.

As an initial model, we train a single decision tree using the 20 prosodic features listed in Table 1. Using a 20-fold leave-one-speaker-out cross-validation approach to evaluate this model over all the utterances in our corpus, we find that it classifies self-reports with an accuracy of 66.33%. As shown in Table 2, the single decision tree model does better than the naive baseline of choosing the most-common class, which has an accuracy of 52.30%, and marginally better than assigning the self-reported certainty to be the same as the perceived certainty, which has an accuracy of 63.67%. Still, we would like to know if we could do better than 66.33%.

As an alternative approach, suppose we know an utterance's perceived level of certainty. Could we use this knowledge, along with the prosody of the utterance to better predict the self-reported certainty? To test this, we divide the data into four subsets (see Figure 4) corresponding to the correctness of the answer and the perceived level of certainty.

In subset $A'$, the distribution of self-reports is heavily skewed; 84% of the utterances in subset $A'$ are self-reported as *uncertain*. This imbalance is intuitive; someone who is incorrect *and* perceived as uncertain most likely feels uncertain too. Likewise, in subset $B'$, the distribution of self-reports is skewed in the other direction; 76% of the utterances in this subset are self-reported as *certain*. This too is intuitive; someone who is correct *and* perceived as certain most likely feels certain as well. Therefore, we hypothesize that for subsets $A'$ and $B'$, classification models trained on prosodic features will do no better than choosing the subset-specific majority class.

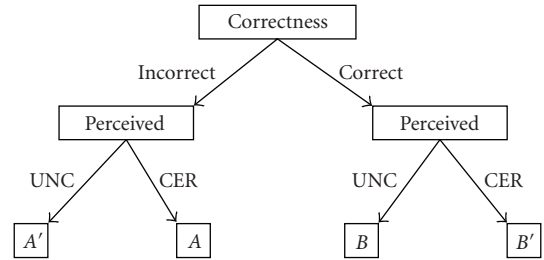

FIGURE 4: We divide the utterances into four subsets and train a separate classifier for each subset.

TABLE 3: Accuracies for classifying self-reported level of certainty for the prosodic decision tree models trained separately on each of the four subsets of utterances. For subsets $A$ and $B$, the decision trees perform better than assigning the subset-majority class, while for subsets $A'$ and $B'$, the decision trees do no better than assigning the subset-majority class. The combined decision tree model has an overall accuracy of 75.30%, significantly better than the single-decision tree (66.33%).

| Subset | Accuracy (subset majority) | Accuracy (prosody decision tree) |
|---|---|---|
| $A$ | 65.19 | **68.99** |
| $B$ | 53.52 | **69.01** |
| $A'$ | 84.35 | 84.35 |
| $B'$ | 75.89 | 75.89 |
| Overall | 72.49 | **75.30** |

Subsets $A$ and $B$ are the more interesting cases; they are the subsets where the perceived level of certainty is not aligned with the correctness. The self-reported levels of certainty for these subsets are less skewed: 65% *uncertain* for subset $A$ and 54% *certain* for subset $B$. We hypothesize that for subsets $A$ and $B$, decision tree models trained on prosodic features will be more accurate than selecting the subset-specific majority class. For each subset, we perform a $k$-fold cross-validation, where we leave one speaker out of each fold. Because not all speakers have utterances in every subset, $k$ ranges from 18 to 20.

*4.3. Results.* For subset $A$, the decision tree accuracy in classifying the self-reported level of certainty is 68.99%, while assigning the subset-majority class (uncertain) results in an accuracy of 65.19%. For subset $B$, the decision tree accuracy is 69.01%, while assigning the subset-majority class (certain) results in an accuracy of 53.52%. Thus, for these two subsets, the prosody of the utterance is more informative than the majority-class baseline. As expected, for subsets $A'$ and $B'$, the decision tree models do no better than assigning the subset-majority class. These results are summarized in Table 3.

The combined decision tree model has an overall accuracy of 75.30%, significantly better than the single-decision tree model (66.33%), which assumed no knowledge of the correctness or the perceived level of certainty. Our combined decision tree model also outperforms the decision tree that has knowledge of prosody and of correctness but lacks knowledge of the perceived certainty; this tree ignores the prosody and splits only on correctness (72.49%). Therefore, if we know an utterance's perceived level of certainty, we can use that information to much more accurately model the self-reported level of certainty.

## 5. Modeling Perceived Level of Certainty

We saw in Section 4 that knowing whether an utterance was perceived as certain or uncertain allows us to make better predictions about the speaker's actual level of certainty. Furthermore, as discussed in Section 1, perceived level of certainty is in and of itself useful in dialogue applications. So, we would like to have a model that tells us how certain a person *sounds* to an average listener, which we turn to now.

*5.1. Basic Prosody Model.* For the basic model, we compute values for the 20 prosodic features listed in Table 1 for each utterance in the corpus. We use these features as input variables to a simple linear regression model for predicting perceived level of certainty scores (on the 1 to 5 scale). To evaluate our model, we divide the data into 20 folds (one fold per speaker) and perform a 20-fold cross-validation. That is, we fit a model using data from 19 speakers and test on the remaining speaker. Thus, when we test our models, we are testing the ability to classify utterances of an unseen speaker.

*5.2. Nonprosodic Model.* We want to ensure that the predictions our prosodic models make are not able to be explained by nonprosodic features such as a word's length, familiarity, or part of speech, or an utterance's position in the data collection materials. Therefore, we train a linear regression model on a set of nonprosodic features to serve as a baseline.

Our nonprosodic model has 20 features. Many of these features assume knowledge of the utterance's target word, the word or phrase that is the probable source of uncertainty. Because the basic prosody model does not assume knowledge of the target word, we consider this to be a generous baseline for this experiment. (In Section 6, we present a prosody model that *does* assume knowledge of the target word.)

The part-of-speech features include binary features for the possible parts of speech of the target word and of its immediately preceding word. Utterance position is represented as the utterance's ordinal position among the sequence of items. (The order varied for each speaker.) Word position features include the target word's index from the start of the utterance, index from the end, and relative position (index from start/total words in utterance). The word length features include the number of characters, phonemes, and syllables in the target word. To account for familiarity, we include a feature for how many times during the experiment the speaker has previously uttered the target word. To approximate word frequency, we use the log probability based on British National Corpus counts where available. For words that do not appear in the British National corpus, we estimate feature values by using web-based counts (Google hits) to interpolate unigram frequencies. It has been demonstrated that using web-based counts is a reliable method for estimating unseen $n$-gram frequencies [17].

Table 4: Our basic prosody model uses utterance-level prosodic features to fit a linear regression (LR) model. This set of input variables performs significantly better than a linear regression model trained on nonprosodic features, as well as the naive baseline of choosing the most common class. The improvement over this naive baseline is on par with prior work.

|  | RMS error (LR model) | Accuracy (LR model) | Accuracy (prior work) |
|---|---|---|---|
| Naive baseline | — | 56.25 | 66.00 |
| Nonprosodic baseline | 1.059 | 51.00 | — |
| Utterance-level prosody | 0.738 | 68.96 | 76.42 |

*5.3. Results.* Since our basic prosody model and our nonprosodic baseline model are linear regression models, comparing root-mean-squared (RMS) error of the two models tells us how well they fit the data. We find that our basic prosody model has lower RMS error than the nonprosodic baseline model: 0.738 compared to 1.059. Table 4 shows the results comparing our basic prosody model against the nonprosodic mdoel.

We also compare our basic prosody model to the prior work of Liscombe et al. [8], whose prosodic input variables are similar to our basic model's input variables. However, we note that our evaluation is more rigorous. While we test our model using a leave-one-speaker-out cross-validation approach, Liscombe et al. randomly divide their data into training and test sets, so that the test data includes utterances from speakers in the training set, and they run only a single split, so their reported accuracy may not be indicative of the entire data set.

Our model outputs a real-valued score; the model of Liscombe et al. [8] outputs one of three classes: certain, uncertain, or neutral. To compare our model against theirs, we convert our scores into three classes by first rounding to the nearest integer, and then coding 1 and 2 as uncertain, 3 as neutral, and 4 and 5 as certain. (This partition of the 1–5 scores is the one that maximizes interannotator agreement among the five human judges.) Table 4 shows the results comparing our basic prosody model against this prior work.

Liscombe et al. [8] compare their model against the naive baseline of choosing the most common class. For their corpus, this baseline was 66.00%. In our corpus, choosing the most-common class gives an accuracy of 56.25%. Our model's classification accuracy is 68.96%, a 12.71% difference from the naive baseline, corresponding to a 29.05% reduction in error. Liscombe et al. [8] report an accuracy of 76.42%, a 10.42% difference from the naive baseline, and 30.65% reduction in error. Thus, our basic model's improvement over the naive baseline is on par with prior work.

In summary, our basic prosody model, which uses utterance-level prosodic features as input, performs better than a substantive nonprosodic baseline model, and also better than a naive baseline model (choosing the majority class), on par with the classification results of prior work.

## 6. Feature Selection for Modeling Perceived Level of Certainty

In the previous section, we showed that our basic prosody model performs better than two baseline models and on par with previous work. In this section, we show how to improve upon our basic prosody model through context-based feature selection. Because the nature of our corpus (see Section 2) makes it possible to isolate a single word or phrase responsible for variations in a speaker's level of certainty, we have good reason to consider using prosodic features not only at the utterance level, but also at the word and phrase level.

*6.1. Utterance, Context, and Target Word Prosodic Features.* For each utterance, we compute three values for each of the 20 prosodic features listed in Table 1: one value for the whole utterance, one for the context segment, and one for the target word segment, resulting in a total of 60 prosodic features per utterance. Target word segmentation was done manually; pauses are considered part of the word that they precede. While the 20 features listed in Table 1 are comparable to those used in previous uncertainty classification experiments [8], to our knowledge, no previous work has used features extracted from context or target word segments.

*6.2. Correlations.* To aid our feature selection decisions, we examine the correlations between an utterance's perceived level of certainty and the 60 prosodic features described in Section 6.1. Correlations are reported for 480 of the 600 utterances in the corpus, those which contain exactly one target word. (Some of the items had two or three slots for target words.) The correlation coefficients are reported in Table 5. While some prosodic cues to level of certainty, such as *total silence*, are strongest in the whole utterance, others are stronger in the context or the target word segments, such as *range f0* and *speaking rate*. These results suggest that models trained on prosodic features of the context and target word may be better than those trained on only whole utterance features.

*6.3. Feature Sets.* We build linear regression level-of-certainty classifiers in the same way as our basic prosody model, only now we consider different sets of prosodic input features. We call the set of 20 whole utterance features from the basic model set A. Set B contains only target word features. Set C contains only context features. Set D is the union of A, B, and C. And lastly, set E is the "combination" feature set—a set of 20 features that we designed based on our correlation results (see Section 6.2). For each prosodic feature type (i.e., each row in Table 5), we select either the whole utterance feature, the context feature, or the target word feature, whichever one has the strongest correlation with perceived level of certainty. The features comprising the combination set are listed below.

Table 5: Correlations between mean perceived rating and prosodic features for whole utterances, contexts, and target words, $N = 480$ (note: *indicates significant at $P < .05$; **indicates significant at $P < .01$).

| Feature | Whole utterance | Context | Target word |
|---|---|---|---|
| Min f0 | 0.107* | 0.119* | 0.041** |
| Max f0 | −0.073 | −0.153** | −0.045 |
| Mean f0 | 0.033 | 0.070 | −0.004 |
| Stdev f0 | −0.035 | −0.047 | −0.043 |
| Range f0 | −0.128** | −0.211** | −0.075 |
| Rel. position min f0 | 0.042 | 0.022 | 0.046 |
| Rel. position max f0 | 0.015 | 0.008 | 0.001 |
| Absolute slope f0 | 0.275** | 0.180** | 0.191** |
| Min RMS | 0.101* | 0.172** | 0.027 |
| Max RMS | −0.091* | −0.110* | −0.034 |
| Mean RMS | −0.012 | 0.039 | −0.031 |
| Stdev RMS | −0.002 | −0.003 | −0.019 |
| Rel. position min RMS | 0.101* | 0.172** | 0.027 |
| Rel. position max RMS | −0.039 | −0.028 | −0.007 |
| Total silence | −0.643** | −0.507** | −0.495** |
| Percent silence | −0.455** | −0.225** | −0.532** |
| Total duration | −0.592** | −0.502** | −0.590** |
| Speaking duration | −0.430** | −0.390** | −0.386** |
| Speaking rate | 0.090* | 0.014 | 0.136** |

(1) *Whole Utterance*. total silence, total duration, speaking duration, relative position max f0, relative position max RMS, absolute slope (Hz), and absolute slope (semitones).

(2) *Context*. min f0, max f0, mean f0, stdev f0, range f0, min RMS, max RMS, mean RMS, and relative position min RMS.

(3) *Target Word*. percent silence, speaking rate, relative position min f0, and stdev RMS.

For each set of input features, we evaluate the model by dividing the data into 20 folds and performing a leave-one-speaker-out cross-validation, as in Section 5.1.

*6.4. Results.* Table 6 shows the accuracies of the models trained on the five subsets of features. The numbers reported are averages of the 20 cross-validation accuracies. To compare these results with those in Table 4, we convert the linear regression output to certain, uncertain, and neutral classes, as described in Section 5.3. As before, the naive baseline is the accuracy that would be achieved by always choosing the most common class, and the nonprosodic baseline model is the same as described in Section 5.2.

Table 6: Average classification accuracies for the linear regression models trained on five subsets of prosodic features. The model trained on the combination feature set performs significantly better than the utterance, target word, and context feature sets.

| Feature set | Num. features | Accuracy |
| --- | --- | --- |
| Naive baseline | N/A | 56.25 |
| Nonprosodic baseline | 20 | 51.00 |
| (A) Utterance | 20 | 68.96 |
| (B) Target word | 20 | 68.96 |
| (C) Context | 20 | 67.50 |
| (D) All | 60 | 74.58 |
| (E) Combination | 20 | 74.79 |

*6.5. Discussion.* The key comparison to notice is that the combination feature set E, with only 20 features, yields higher average accuracies than the utterance feature set A: a difference of 5.83%. This suggests that using a combination of features from the context and target word in addition to features from the whole utterance leads to better prediction of the perceived level of certainty than using features from only the whole utterance.

One might argue that these differences are just due to noise. To address this issue, we compare the prediction accuracies of sets A and E per fold. Each fold in our cross-validation corresponds to a different speaker, so the folds are *not* identically distributed, and we do not expect each fold to yield the same prediction accuracy. That means that we should compare predictions of the two feature sets within folds rather than between folds. Figure 5 shows the correlations between the predicted and perceived levels of certainty for the models trained on sets A and E. The combination set E predictions were more strongly correlated than whole utterance set A predictions in 16 out of 20 folds. This result supports our claim that using a combination of features from the context and target word in addition to features from the whole utterance leads to better prediction of level of certainty.

Figure 5 also shows that one speaker (the 17th fold) is an outlier—for this speaker, our model's level of certainty predictions are less correlated with the perceived levels of certainty than for all other speakers. Most likely, this results from nonprosodic cues of uncertainty present in the utterances of this speaker (e.g., disfluencies). Removing this speaker from our training data did not improve the overall performance of our models.

These results suggest a better predictive model of level of certainty for systems where words or phrases likely to cause uncertainty are known ahead of time. Without increasing the total number of features, combining select prosodic features from the target word, the surrounding context and the whole utterance lead to better prediction of level of certainty than using features from the whole utterance only.

## 7. Detecting Uncertainty at the Phrase Level

In Section 6, we showed that incorporating the prosody of the target word and of its context into our level of certainty models improves classification accuracy. In this section, we show that our models can be used to make an informed guess about which phrase a person is uncertain about, when we do not know which phrase is the probable source of uncertainty. As an initial step towards the problem of identifying one phrase out of all possible phrases, we ask a simpler question: given two phrases, one that the speaker is uncertain about (the target word), and another phrase that they are not uncertain about (a control word), can our models determine which phrase is causing the uncertainty? Using the prosody-based level-of-certainty classification models described in Section 6.3, we compare the predicted level of certainty using the actual target word segmentation with the predicted level using an alternative segmentation with a control word as the proposed target word. Our best model is able to identify the correct segmentation 91% of the time, a 71% error reduction over the baseline model trained on only nonprosodic features.

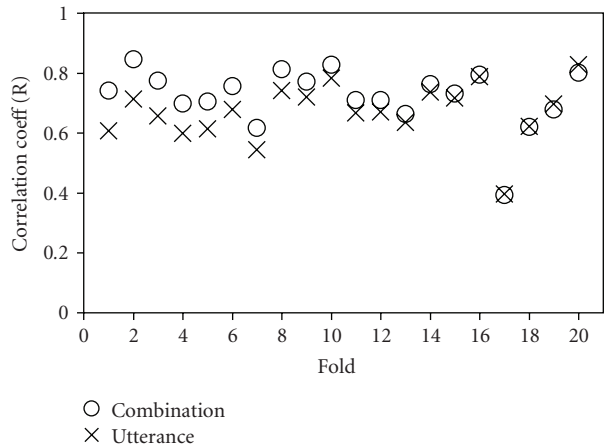

Figure 5: Correlations with perceived level of certainty per fold for the combination (O) and the utterance (X) feature set predictions, sorted by the size of the difference. In 16 of the 20 experiments, the correlation coefficients for the combination feature set are greater than those of the utterance feature set.

*7.1. Experiment Design.* For a subset of utterances that were perceived to be uncertain (perceived level of certainty less than 2.5), we identify a control word—a content word roughly the same length as the potential target words and if possible, the same part of speech. In the example item shown below, the control word used was *abrasive*.

> Mahler's revolutionary music, abrasive personality, and ____ writings about art and life divided the city into warring factions.
>
> (a) officious
> (b) trenchant
> (c) spoffish
> (d) pugnacious

We balance the set of control words for position in the utterance relative to the position of the slot; half of the control words appear before the slot location and half appear after. After filtering utterances based on level of certainty and presence of an appropriate control word, 43 utterances remain. This is our test set.

Table 7: Accuracies on the task of identifying the word or phrase causing uncertainty when choosing between the actual word and a control word. The model that was trained on the set of target word features and nonprosodic features achieves 91% accuracy.

| Feature set | Num. features | Detection accuracy |
| --- | --- | --- |
| Nonprosodic baseline | 20 | 67.44 |
| Target word, nonprosodic | 40 | 90.70 |
| Target word | 20 | 86.05 |
| Target word, context, utterance | 60 | 79.07 |
| Target word, context, utterance, nonprosodic | 80 | 76.74 |
| Target word, utterance | 40 | 69.77 |
| Combination set (target word) | 4 | 72.09 |
| Combination set (target word, context, utterance) | 20 | 72.09 |
| Context | 20 | 48.84 |

We then compare the predicted level of certainty for two segmentations of the utterance: (a) the correct segmentation with the slot-filling word as the proposed "target word" and (b) an alternative segmentation with the control word as the proposed "target word." Thus, the prosodic features extracted from the target word and from the context will be different in these two segmentations, while the features extracted from the utterance will be the same. The hypothesis we test in this experiment is that our models should predict a lower level of certainty when the prosodic features are taken from segmentation (a) rather than segmentation (b), thereby identifying the slot-filling word as the source of the speaker's uncertainty.

Our models are the same ones described in Section 6.3. They are trained on the same 60 prosodic features from each whole utterance, context, and target word (see Section 6.1) and evaluated with a leave-one-speaker-out cross-validation as before. We use the nonprosodic model described in Section 5.2 as a baseline for this experiment.

*7.2. Results.* Our models yield accuracies as high as 91% on the task of identifying the word or phrase causing uncertainty when choosing between the actual word and a control word. Table 7 shows the linear regression accuracies for a variety of feature sets. The models trained on the nonprosodic features provide a baseline from which to compare the performance of the models trained on prosodic features. This baseline accuracy is 67%.

The linear regression model trained on the target word feature set had the highest accuracy among the purely prosodic models, 86%. The highest overall accuracy, 91%, was achieved on the model trained on the target word features plus the nonprosodic features from the baseline set. We also trained support vector machine models using the same feature sets. The accuracy of these models was on par with or lower than the linear regression models [18].

*7.3. Discussion.* This experiment shows that prosodic level-of-certainty models are useful in detecting uncertainty at the word level. Our best model, the one that uses target word prosodic features plus the nonprosodic features from the baseline set identifies the correct word 91% of the time whereas the baseline model using only nonprosodic features is accurate just 67% of the time. This is an absolute difference of 23% and an error reduction of 71%. This large improvement over the nonprosodic baseline model implies that prosodic features are crucial in word-level uncertainty detection.

In creating the nonprosodic feature set for this experiment, we wanted to account for the most obvious differences between the target words and the control words. The baseline model's low accuracy on this task is to be expected because the nonprosodic features are not good at explaining the variance in the response variable (perceived level of certainty): the correlation coefficient for the baseline linear regression model is only 0.27. (As a comparison, the coefficient for the target word linear regression model is 0.67.)

The combination feature set, which had high accuracy in classifying an utterance's overall level of certainty, did not perform as well as the other feature sets for this detection task. We speculate that this may have to do with the context features. While the prosodic features we extracted from the context are beneficial in classifying an utterance's overall level of certainty, the low accuracies for the context feature set in Table 7 suggest that they are detrimental in determining which word a speaker is uncertain about, using our proposed method. The task we examine in this section, distinguishing the actual target word from a control word, is different from the task the models are trained on (predicting a real-valued level of certainty); therefore, we do not expect the models with the highest classification accuracy to necessarily perform well on the task of identifying the word causing uncertainty.

## 8. Conclusion

Imagine a computer tutor that engages in conversation with a student about particular topics. Adapting the tutor's future behaviors based on knowledge of whether the student is confident in his or her responses could benefit both the student's learning gain and satisfaction with the interaction [2]. A student's response to a question, incorporating language from the question augmented by a student-generated phrase or two, incorporates phrase-level prosodic information that provides clues to the internal level of certainty of the student. The results we have presented provide some first indications that knowledge of which phrases were likely to have engendered uncertainty can significantly enhance the system's ability to predict level of certainty, and even to select which phrase is the source of uncertainty.

Overall, our results suggest that we can get a good estimate of a speaker's level of certainty based on only prosodic features. In our experiments, we used a small set of the many possible prosodic features that have been examined in related work. Because these features proved beneficial in recognizing uncertainty, we believe that using an expanded set of prosodic features might be even more beneficial. In natural conversation, people also convey uncertainty through other channels such as body language, facial gestures, and word choice. Further work is needed to understand how to integrate cues from multiple modalities, when these other modes of input are available.

Our results were enabled by a novel methodology for collecting uncertainty data that allowed us to isolate the phrase causing uncertainty. We also addressed a question that is important to all research regarding mental or emotional state modeling—the difference between a person's self-reported state and an outsider's perception of that state. In our corpus, these two quantities are aligned for approximately one-half of the utterances and mismatched for the remaining half, suggesting that classifiers trained on only perceived judgements of certainty may end up missing actual instances of uncertainty. This highlights the importance of collecting data in ways that maximize our ability to externally control or ensure access to a person's internal mental state. It also raises the question of whether computers may even surpass humans at classifying a speaker's internal level of certainty.

## Acknowledgment

This work was supported in part by a National Defense Science and Engineering Graduate Fellowship.